\documentclass{article}
\usepackage{spconf,amsmath,graphicx}
\usepackage{paralist,amssymb, amsfonts}
\DeclareMathOperator*{\argmax}{arg\,max}
\DeclareMathOperator*{\argmin}{arg\,min}

\title{Learning fashion compatibility across apparel categories for outfit recommendation}
\name{Luisa~F.~Polan\'{i}a, Satyajit Gupte}
\address{Target Corporation, EDABI, Sunnyvale, CA}

\begin{document}
%
\maketitle
\begin{abstract}
This paper addresses the problem of generating recommendations for completing the outfit given that a user is interested in a particular apparel item. The proposed method is based on a siamese network used for feature extraction followed by a fully-connected network used for learning a fashion compatibility metric. The embeddings generated by the siamese network are augmented with color histogram features motivated by the important role that color plays in determining fashion compatibility. The training of the network is formulated as a maximum a posteriori (MAP) problem where Laplacian distributions are assumed for the filters of the siamese network to promote sparsity and matrix-variate normal distributions are assumed for the weights of the metric network to efficiently exploit correlations between the input units of each fully-connected layer. 


\end{abstract}
\section{Introduction}
\label{sec:intro}

New vision problems are arising in response to the rapid evolution of the online fashion industry~\cite{Dong17, Saha18, Chen17, Rubi17}. In recent years, the problem of predicting fashion compatibility for outfit recommendation has gained popularity in the vision community \cite{Mcau15, He16, Chen18, Hsia18, Song18, Zhan18, Li17}. This problem goes beyond the traditional problem of visual similarity because it requires modeling and inferring the compatibility relations across different fashion categories, as well as the relations between multiple fashion factors, such as color, material, pattern, texture, and shape. It is also highly subjective because fashion compatibility might vary from one person to another, which may lead to noisy labels.

Fashion compatibility can be posed as a metric learning problem \cite{Mcau15, He16}, addressable with Siamese networks. In \cite{Mcau15}, McAuley \textit{et al}. used parameterized distance metric to learn relationships between co-purchased item pairs and used convolution neural networks (CNNs) for feature extraction. Chen \textit{et al}. \cite{Chen18} proposed a triplet loss-based metric learning method to recommend complementary fashion items. An alternative approach to metric learning is to use recurrent neural networks \cite{Han17} to model outfit generation as a sequential process. Hsiao \textit{et al}. \cite{Hsia18} proposed to create capsule wardrobes from fashion images by posing the task as a subset selection problem. Song \textit{et al}. \cite{Song18} proposed to model compatibility  using an attentive knowledge distillation scheme. Vasileva \textit{et al}. \cite{Vasi18} trained a network to learn similarity and compatibility simultaneously in different spaces for each pair of item categories.





Given that a customer is interested in an apparel item, referred to as query, the method proposed in this paper recommends complementary apparel items that matches the query to form  a stylish outfit. The proposed method uses siamese networks for feature extraction. The methods in \cite{Veit15} and \cite{Mcau15} also use siamese networks, but they simply use the Euclidean and Mahalanobis distances as the compatibility score, in contrast to our work, which calculates the compatibility score using a fully-connected (FC) network, which has the potential to generate more complex compatibility functions beyond distance metrics such as Euclidean or Mahalanobis. The simplest approach to merge the image embeddings generated by the siamese network is to use concatenation. Instead, the proposed method uses the Hadamard product to directly encode correlation between the embeddings and therefore, help the network learn a compatibility metric.

Another contribution of this paper is that it explicitly incorporates color information into the network, and therefore, does not completely rely on the CNN to learn the color features that are relevant for fashion compatibility. It has been shown that color plays a critical role in determining compatibility between fashion items \cite{Jaga14}. 

In terms of training, we adopt a similar approach to that of \cite{Zhan15}, where a MAP approach is employed. Unlike \cite{Zhan15}, a Laplacian distribution, which promotes sparsity, is used to model the CNN filters. In addition to the benefits of less storage and faster inference, sparsity-promoting priors also lead to redundancy reduction in the weights. In \cite{Zhan15}, the matrix-variate normal distribution with unit row and unit column normalized covariance matrices is assumed for the weights of the FC layer. In contrast, the proposed approach does not impose that constraint on the column covariance of the distribution and therefore, is able to effectively capture the correlations between the input units of each FC layer.




\section{Method}
\label{sec:intro}
The proposed network is composed of two sub-networks, a siamese sub-network and a metric learning sub-network. The sub-networks are jointly learned in an end-to-end fashion. The siamese sub-network maps a pair of apparel images to a pair of features and the metric sub-network maps the pair of features and auxiliary cues to a fashion compatibility score.

\begin{figure}[t]
\centering{ 
\includegraphics[width =.78\columnwidth]{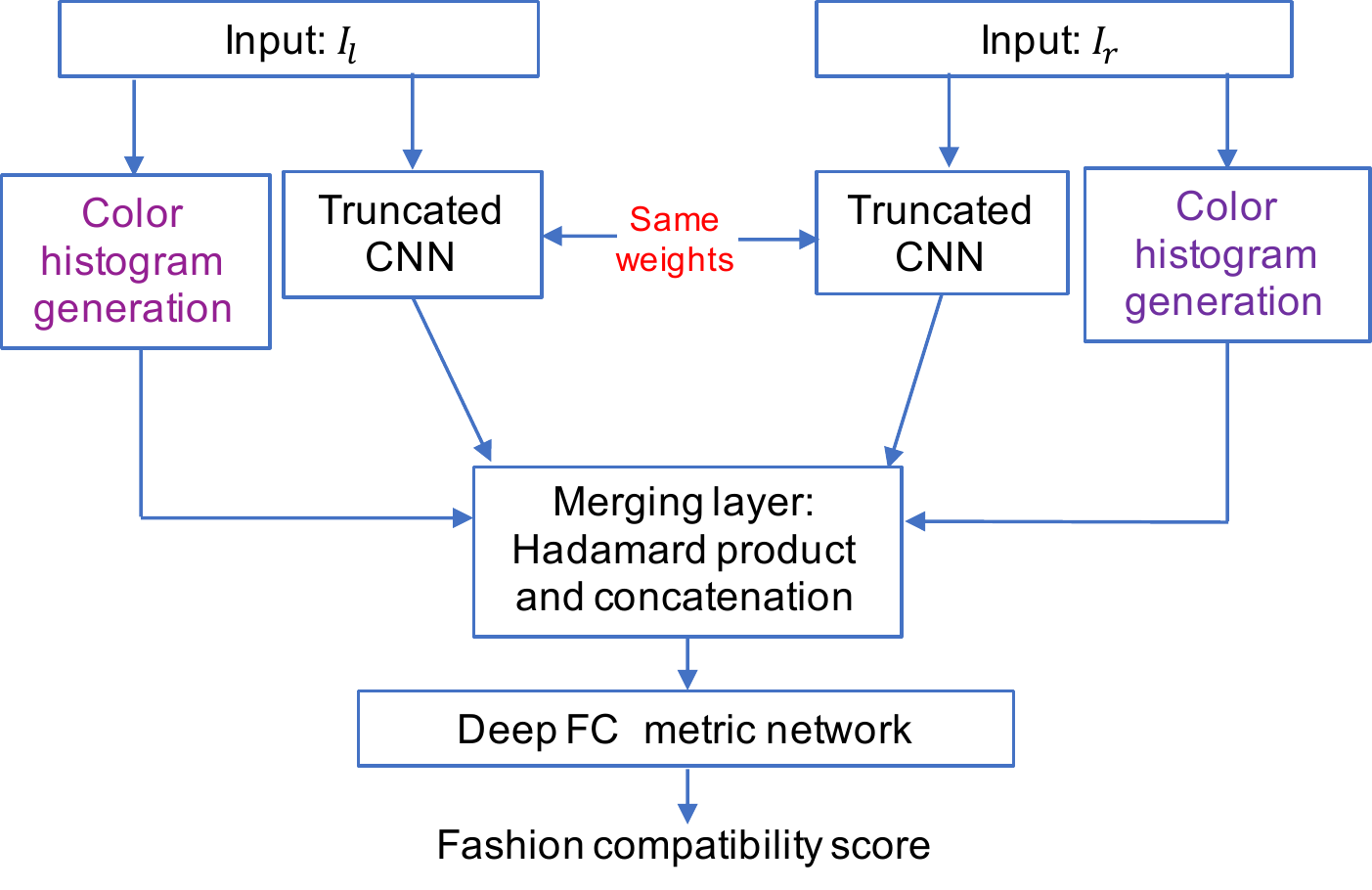}}
\caption{Schematic of the proposed architecture.} \label{polyvore}
\end{figure}

Siamese networks are composed of two branches with shared weights. In this paper, the branches are two identical truncated VGG-16 nets \cite{Simo14} pre-trained on ImageNet. The nets are truncated in the sense that the FC layers are excluded. The embeddings generated by the siamese sub-network are merged using the Hadamard product to directly encode the correlation between embeddings.  This merging strategy simulates the adaptive weighted cross-correlation technique \cite{Heo11}. The Hadamard product of the color histogram features extracted from the pair of input images is also calculated and concatenated with the Hadamard product of the siamese embeddings. The resulting concatenated vector is forward-propagated through a FC metric sub-network and a readout function, which is a simple linear regression, to compute the targets. A schematic of the proposed architecture is shown in Fig. \ref{polyvore}.

The $N$ training input image pairs are denoted as $\mathbf{I}=\{(I_r, I_l)_i\}_{i=1}^N$, where $I_r$ and $I_l$ denote the inputs to the right and left branches of the siamese sub-network, respectively. Let $\mathbf{Y}=\{y_i\}_{i=1}^N$ denote the binary labels, where $y_i$ is 1 if the input pair $(I_r, I_l)_i$ is fashion compatible and 0 otherwise. Let $x_i$ be the output of the last FC layer of the metric sub-network when the pair $(I_r, I_l)_i$ is used as input. The readout function $\Gamma(\cdot)$ takes the form
\begin{equation} \label{readout}
\hat{y}_i=\Gamma(x_i)=w^Tx_i+\epsilon,
\end{equation}
where $w$ are the weights of the readout function and $\epsilon$ has a standard logistic distribution. Batch normalization layers are applied after each FC layer of the metric sub-network. 

Even though the siamese sub-network has two sets of weights, $\mathbf{\Theta}_r$ and $\mathbf{\Theta}_l$ for the right and left branches, respectively, they are mirrored, and therefore, simply referred to as $\mathbf{\Theta}$, \textit{i.e.}, $\mathbf{\Theta}=\mathbf{\Theta}_r=\mathbf{\Theta}_l$. Let $\mathbf{\Theta}_s=\{\theta_t\}_{t=1}^{S}$ be a subset of $\mathbf{\Theta}$ that corresponds to the $S$ filters selected for fine-tuning. Let $\mathbf{W}=\{W_j \in \mathbb{R}^{P_j\times Q_j} \}_{j=1}^M$ be the set of weights of the FC metric sub-network. Those weights are modeled with a matrix-variate normal distribution of zero mean, \textit{i.e.} $W_j \sim \mathcal{MN}(\mathbf{0}, \Lambda_j, \gamma_j^2\mathcal{I}), \forall j$, where $\mathbf{0} \in \mathbb{R}^{P_j\times Q_j}$ is a zero matrix, $\gamma_j^2\mathcal{I} \in \mathbb{R}^{P_j\times P_j}$, the row covariance, is a diagonal matrix with diagonal elements $\gamma_j^2$, and $\Lambda_j \in \mathbb{R}^{Q_j\times Q_j}$ is the positive semi-definite column covariance matrix, which can be learned in order to capture correlations between the layer input units. Let $\mathbf{\Lambda}=\{\Lambda_j\}_{j=1}^M$ be the set of column covariance matrices.

The proposed network is trained by solving the following MAP problem

\begin{eqnarray}
\hat{\mathbf{\Theta}}_s, \hat{\mathbf{W}}, \hat{\mathbf{\Lambda}}, \hat{w}&=&\argmax_{{\mathbf{\Theta}}_s, {\mathbf{W}}, {\mathbf{\Lambda}}, {w}} p({\mathbf{\Theta}}_s, {\mathbf{W}}, {\mathbf{\Lambda}}, {w}|\mathbf{I}, \mathbf{Y}),\\ \label{map}
&=& \argmax_{{\mathbf{\Theta}}_s, {\mathbf{W}}, {\mathbf{\Lambda}}, {w}} p(\mathbf{Y}|\mathbf{I}, \mathbf{\Theta}_s, \mathbf{W}, w)\times \\ 
&&p(\mathbf{W}|\mathbf{\Lambda})\times p(\mathbf{\Theta}_s) \times p(w) .  \nonumber
\end{eqnarray}
The first term in \eqref{map} is the likelihood probability while the last three terms are prior probabilities. Given that the noise in \eqref{readout} has a logistic distribution, the probability distribution of $y_i$ given $x_i$ is Bernoulli, and therefore, the likelihood is
\vspace{-.05cm}
\begin{eqnarray} \label{readout2}
p(\mathbf{Y}|\mathbf{I}, \mathbf{\Theta}_s, \mathbf{W}, w)\propto \quad \quad \quad \quad \quad \quad \quad \quad \quad \quad \quad 
\\
\prod_{i=1}^N p(y_i=1|x_i)^{y_i}(1-p(y_i=1|x_i))^{(1-y_i)}, \nonumber
\end{eqnarray}
where $p(y_i=1|x_i)=r(w^Tx_i)=1/(1+\text{exp}\{-w^Tx_i\})$, where $r(\cdot)$ is the sigmoid function.

Each entry of the vectors in the set $\mathbf{\Theta}_s$ and each entry of $w$ is modeled with a Laplacian distribution of zero mean and variance $\sigma_i^2$, in the case of $\theta_i$, and variance $\sigma_w^2$, in the case of $w$. The motivation for using a Laplacian distribution is to promote sparsity, and therefore, reduce redundancy in the weights. Since $W_i$ is modeled  by a matrix-variate normal distribution with zero mean, the prior probability $p(\mathbf{W}|\mathbf{\Lambda})$ takes the form
\vspace{-.1cm}
\begin{eqnarray}
p(\mathbf{W}|\mathbf{\Lambda}) &=& \prod_{j=1}^M p(W_j|\Lambda_j)\\
&=& \prod_{j=1}^M \frac{\exp \left(-\frac{1}{2} \text{tr}{\left((\gamma_j^2I)^{-1}W_j\Lambda_j^{-1}W_j^T\right)}  \right)}{(2\pi)^{\frac{P_jQ_j}{2}} {|\gamma_j^2I|}^{\frac{P_j}{2}}  {|\Lambda_j|}^{\frac{Q_j}{2}}  }, \label{norm_matrix}
\end{eqnarray}
where $\text{tr}(\cdot)$ and $|\cdot|$ denote the trace and determinant of a matrix, respectively.

Replacing equations \eqref{readout2}, \eqref{norm_matrix} and the Laplacian priors for $w$ and $\mathbf{\Theta}_s$  in \eqref{map},  setting the variance parameters $\sigma_w$, $\{\sigma_i\}_{i=1}^S$, and $\{\gamma_j\}_{j=1}^M$ to 1 for simplicity purposes, removing constant terms, and taking the negative logarithm leads to the following optimization problem
\vspace{-.1cm}
\begin{eqnarray} \label{loss_2}
\hat{\mathbf{\Theta}}_s, \hat{\mathbf{W}}, \hat{\mathbf{\Lambda}}, \hat{w}=\argmin_{{\mathbf{\Theta}}_s, {\mathbf{W}}, {\mathbf{\Lambda}}, {w}}
-\sum_{i=1}^N [y_i \text{ln}(r(w^Tx_i))+ \\ \nonumber
(1-y_i)\text{ln}(1-r(w^Tx_i))] + \sum_{j=1}^M \text{tr}({W}_j\Lambda_j^{-1} {W_j}^T)+\\ \nonumber
\sum_{j=1}^M Q_j\text{ln}|\Lambda_j| + \sum_{t=1}^S \|{\theta}_t\|_1 +  \|w\|_1
\end{eqnarray}
The third summation term in \eqref{loss_2} is concave while the other terms are jointly convex with respect to all variables. The same approach of \cite{Zhan12} of replacing $Q_j\text{ln}|\Lambda_j|$ by the constraint $\text{tr}|\Lambda_j|=1$ is adopted. Therefore, problem \eqref{loss_2} is reformulated as
\vspace{-.25cm}
\begin{eqnarray} \label{loss_3}
\hat{\mathbf{\Theta}}_s, \hat{\mathbf{W}}, \hat{\mathbf{\Lambda}}, \hat{w}=\argmin_{{\mathbf{\Theta}}_s, {\mathbf{W}}, {\mathbf{\Lambda}}, {w}}
-\sum_{i=1}^N [y_i \text{ln}(r(w^Tx_i))+ \\ \nonumber
(1-y_i)\text{ln}(1-r(w^Tx_i))] + \lambda_1\sum_{j=1}^M \text{tr}({W}_j\Lambda_j^{-1} {W_j}^T)+\\ \nonumber
 \lambda_2 \sum_{t=1}^S \|{\theta_t}\|_1 +  \lambda_3 \|w\|_1 ~~\textrm{s.~t.}~~
 {\Lambda_j}\succeq0, \text{tr}|\Lambda_j|=1, \forall j,
\end{eqnarray}
where $\lambda_1, \lambda_2, \lambda_3$ are regularization parameters, which are incorporated to tune the strength of the regularization terms, and are estimated using grid search. The constraint $ {\Lambda_j}\succeq0$ comes from the positive semi-definite property that covariance matrices need to satisfy. An alternating optimization procedure is used to solve for \eqref{loss_3}. First, the filters $\mathbf{\Theta}_s$ and $w$ and the weight matrices $\mathbf{W}$ are updated using stochastic gradient descent while the covariance matrices $\mathbf{\Lambda}$ are kept fixed. Second,  the covariance matrices $\mathbf{\Lambda}$ are updated, while keeping all the other parameters fixed, by using $\hat{\Lambda}_j = \frac{(W_j^TW_j)^{1/2}}{\text{tr} ((W_j^TW_j)^{1/2}) }$ $\forall j$, which, as described in \cite{Zhan12}, is the closed-form solution of 
\begin{equation} \label{trace}
\min_{\Lambda_j} \text{tr}({W}_j\Lambda_j^{-1} {W_j}^T)
 ~~\textrm{s.~t.}~~
 {\Lambda_j}\succeq0, \text{tr}|\Lambda_j|=1.
\end{equation}








\begin{table*}[t]
\footnotesize
\renewcommand{\arraystretch}{1.3}
\caption{Performance comparison using the Lift of Average Precision@K}
\label{table_example}
\centering
\begin{tabular*}{0.68\textwidth}{@{\extracolsep\fill}c|cccccccc}
\hline
Performance metric&{Proposed model}&{M1}&{M2}&{M3}&{M4}&{M5}&{M6 \cite{Veit15}}&{M7 \cite{Vasi18}}\\
\hline
Lift@3&8.67& 3.34&7.09& 3.51 &5.54& 2.9&2.96&5.12\\

Lift@7&5.43&2.01&4.8& 3.38 &3.33&2.08&2.49&3.54\\

Lift@12&4.42&1.61&3.95& 2.89&2.57&1.65&2.03&3.24\\
\hline
\end{tabular*}
\end{table*}

\vspace{-.3cm}
\section{Experimental results}

Training and testing of the proposed approach is performed on a dataset collected from Polyvore, which was a popular fashion website (www.polyvore.com) where users used to create and upload outfit data. We collected our own dataset from Polyvore, which contains 13,947 outfits. These outfits are divided into 10,650, 1,902, 1,395 for training, validation and testing. The outfits are filtered to keep only the apparel categories that fall into the major categories of bottoms, tops, dresses, gowns, suits, and outwear. Outfits whose number of items is less than 2 after filtering are removed from the dataset. Items in the training set do not belong to the testing set and viceversa.


Positive training and validation pairs are built by forming all the possible pair combinations between items belonging to the same outfit. Negative training and validation pairs are built by randomly sampling items from different outfits. Even though this approach has been widely used to generate negative pairs, there is no guarantee that it would lead to true negatives. Therefore, many more negatives than positives are sampled to compensate for the noise in the labels. The oversampling factor is set to 6.

For each outfit in the testing set, the query item is defined as the first item of the outfit. For each of the remaining categories in the outfit, the proposed network generates compatibility scores between the query item and all the items from the collected Polyvore dataset which belong to that category.

The Adam optimizer~\cite{King14} with a base learning rate of $1\times10^{-4}$ and with default momentum values $\beta_1=0.9$ and $\beta_2=0.999$ is used for training with 64 samples per mini-batch. The weights of the first 10 convolutional layers of the truncated VGG-16 net are kept frozen during training. Training stops when the loss on the validation set stops decreasing. ReLU are used as activation functions for the layers in the metric network. For the color histogram, 8 bins are used. The metric network uses 2 FC layers with 256 and 64 hidden units for the first and second layers, respectively. 
\vspace{-.2cm}
\subsection{Methods for comparison and evaluation metrics}
The performance of the proposed network is compared with the following methods:

- M1: Siamese network formed by two identical VGG-16 nets (only convolutional part) pre-trained on ImageNet. Euclidean distance between the siamese embeddings is used to generate the compatibility scores.

-  M2: The proposed network without explicitly incorporating color information in the form of color histograms. 

-  M3: The proposed network but replacing the Hadamard product by concatenation.

- M4: Compatibility score generated by calculating the Euclidean distance between the color histogram of the inputs.

- M5: Compatibility score generated by calculating the Euclidean distance between the HoG features of the inputs (8 orientations and 15$\times$15 pixels per cell are used).

- M6: The siamese architecture proposed in \cite{Veit15}, which uses the Euclidean distance between the learned embeddings. 

- M7: The network proposed in \cite{Vasi18}, which models type-aware compatibility in a conditioned similarity subspace.



\begin{figure}[t]
\centering{ 
\includegraphics[width =.9\columnwidth]{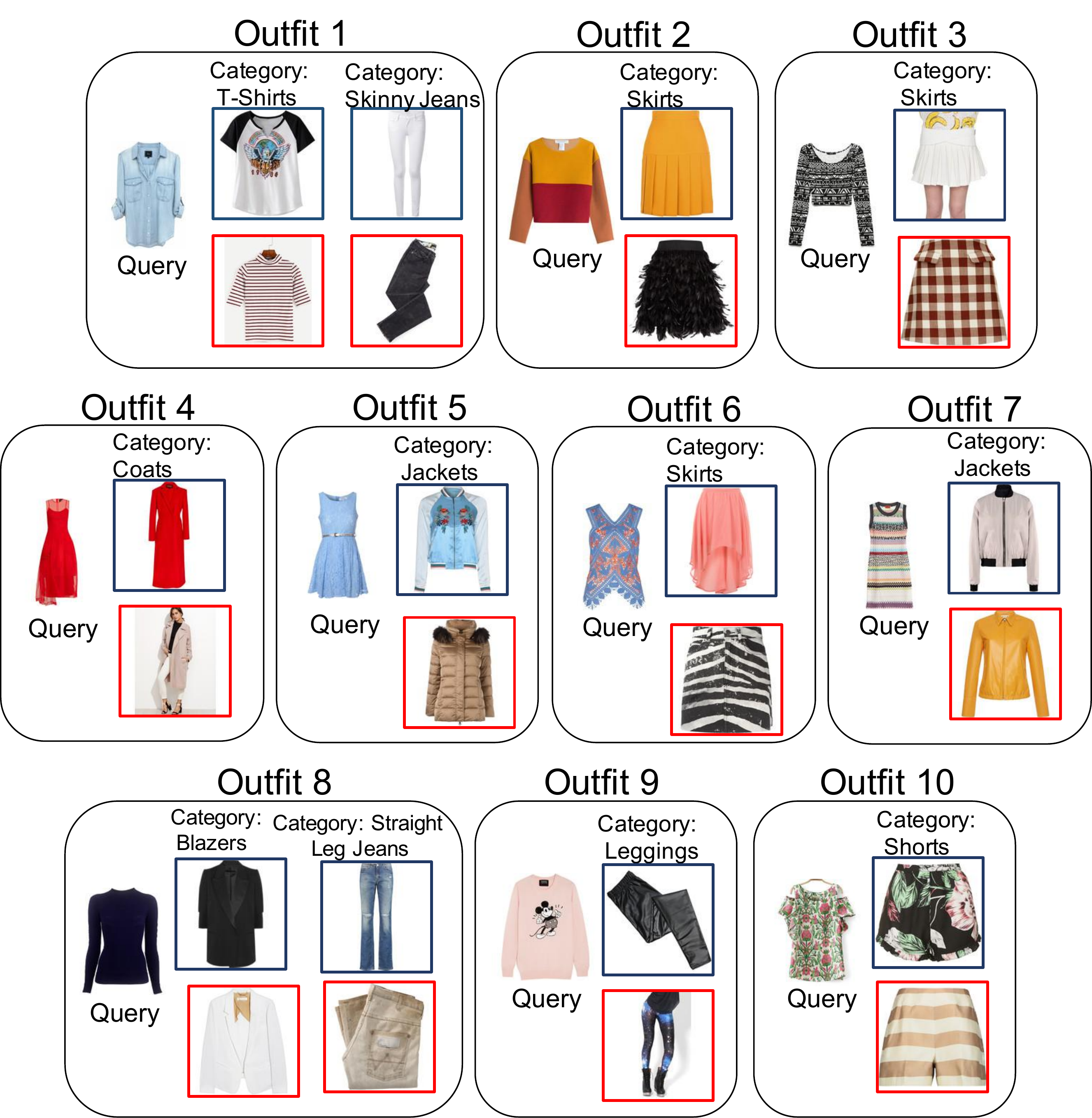}}
\caption{Predictions of the proposed model with the highest and lowest fashion compatibility scores. Images with the highest and lowest scores are outlined in blue and red, respectively.} \label{polyvore2}
\vspace{-0.6cm}
\end{figure}

Let $\{\psi_n\}_{n=1}^{N_t}$ be the set of testing Polyvore outfits, where $\psi_n$ is formed by the query item (first item in the outfit), denoted as $q_n$, and $C_{n}$ complementary items,   denoted as $\{o_n^{(c)}\}_{c=1}^{C_{n}}$, which belong to $C_{n}$ different apparel categories. Let $R_n^c(K)$ denote the top K recommendations generated by the proposed network for the complementary item in category $c$, given the query $q_n$. To generate $R_n^c(K)$, pairs are first formed between the query item and all the rest of the items in the Polyvore dataset which belong to category $c$, then the items with the top $K$ fashion compatibility scores are selected to form $R_n^c(K)$. The precision@$K$ for outfit $\psi_n$ is


\begin{equation} \label{precision}
\text{precision@}K(\psi_n)=\frac{1}{C_n} \sum_{c=1}^{C_n} \mathbf{1}[o_n^{(c)} \in R_n^c(K)],
\end{equation}
where $\mathbf{1}[\cdot]$ denotes the indicator function. The average of the precision@$K$ across the $N_t$ testing outfits is referred to as the average precision@$K$. The recommendation performance of a model is evaluated using the lift of average precision@$K$, which is defined as 
\vspace{-.4cm}

\begin{equation} \label{lift}
\text{Lift@K}=\frac{\text{average precision@}K (\text{model})}{\text{average precision@}K (\text{random})},
\end{equation}
where average precision@$K$ (random) is that of a recommender that would select items at random for each of the apparel categories of interest.

\vspace{-.1cm}

\subsection{Performance Evaluation}
Experimental results are shown in Table \ref{table_example}. By comparing the performance of the proposed model with that of M2, it is clear that explicitly adding color information leads to performance gains, which is probably not surprising since the Lift@$K$, $K=3,7,12$, attained by M4 suggests that using color alone already offers a significant lift. However, recommending items based on the Euclidean distance between color histograms would lead to recommending monochromatic outfits most of the time. The proposed network outperforms that of \cite{Veit15} and \cite{Vasi18}, probably largely due to the explicit incorporation of color features and the metric network, which generates powerful non-linear metric functions. As suggested by the results of M5, HoG features do not play as important role as color features at determining fashion compatibility. 
The comparison of the performance of the proposed model with that of M1 emphasizes the gains attained by jointly fine-tuning the truncated VGG-16 and learning the weights of the metric network. The lift attained by M3 is lower than that of the proposed method, which suggests that the Hadamard product leads to performance gains by efficiently exploiting correlations between the embeddings.

Ten query items are selected from the testing outfits to visually evaluate the model performance. Fig. \ref{polyvore2} illustrates our model predictions for the complementary items with the highest (blue-outlined boxes) and lowest (red-outlined boxes) fashion compatibility scores with respect to the queries. The complementary categories are the same as in the original testing outfit. Results suggest that the network learns color and style relations between apparel categories that lead to stylish outfits to a good extent. Items with the highest compatibility score tend to have either neutral colors, \textit{e.g.} white and black, or to match the colors of the query item. Also, they tend to be either unicolor or contain low-key patterns.  

\vspace{-0.25cm}

\section{Conclusion}
Computer vision is gaining momentum in the fashion industry, where there is huge potential to exploit visual information. In this paper, the proposed method generates recommendations for complementary apparel items given a query item. The proposed method explicitly incorporates color information in the feature extraction process and exploits correlations between the feature representations. The adopted MAP approach used for training promotes sparsity of the weights of the CNN and readout function and allows the metric network to exploit correlations between the input units of the layers.

\vspace{-0.25cm}

\section{Acknowledgements}
The authors thank Kasturi Bhattacharjee for her initial work on this project and for helpful discussions.

\bibliographystyle{IEEEbib}
\bibliography{egbib}

\begin{thebibliography}{10}

\bibitem{Dong17}
Q.~Dong, S.~Gong, and X.~Zhu,
\newblock ``Multi-task curriculum transfer deep learning of clothing
  attributes,''
\newblock in {\em IEEE Winter Conference on Applications of Computer Vision},
  2017, pp. 520--529.

\bibitem{Saha18}
A.~Saha, M.~Nawhal, M.~Khapra, and V.C. Raykar,
\newblock ``Learning disentangled multimodal representations for the fashion
  domain,''
\newblock in {\em IEEE Winter Conference on Applications of Computer Vision},
  2018, pp. 557--566.

\bibitem{Chen17}
Z.~Cheng, X.~Wu, Y.~Liu, and X.~Hua,
\newblock ``Video2shop: Exact matching clothes in videos to online shopping
  images,''
\newblock in {\em IEEE Conference on Computer Vision and Pattern Recognition},
  2017, pp. 4169--4177.

\bibitem{Rubi17}
A.~Rubio, L.~Yu, E.~Simo-Serra, and F.~Moreno-Noguer,
\newblock ``Multi-modal joint embedding for fashion product retrieval,''
\newblock in {\em IEEE International Conference on Image Processing}, 2017, pp.
  400--404.

\bibitem{Mcau15}
J.~McAuley, C.~Targett, Q.~Shi, and A.~Van Den~Hengel,
\newblock ``Image-based recommendations on styles and substitutes,''
\newblock in {\em IACM SIGIR Conference on Research and Development in
  Information Retrieval}, 2015, pp. 43--52.

\bibitem{He16}
R.~He, C.~Packer, and J.~McAuley,
\newblock ``Learning compatibility across categories for heterogeneous item
  recommendation,''
\newblock in {\em International Conference on Data Mining}, 2016, pp. 937--942.

\bibitem{Chen18}
L.~Chen and Y.~He,
\newblock ``Dress fashionably: Learn fashion collocation with deep
  mixed-category metric learning,''
\newblock in {\em AAAI Conference on Artificial Intelligence}, 2018.

\bibitem{Hsia18}
W.~Hsiao and K.~Grauman,
\newblock ``Creating capsule wardrobes from fashion images,''
\newblock in {\em IEEE Conference on Computer Vision and Pattern Recognition},
  2018, pp. 7161--7170.

\bibitem{Song18}
Xuemeng Song, Fuli Feng, Xianjing Han, Xin Yang, Wei Liu, and Liqiang Nie,
\newblock ``Neural compatibility modeling with attentive knowledge
  distillation,''
\newblock {\em arXiv preprint arXiv:1805.00313}, 2018.

\bibitem{Zhan18}
H.~Zhang, W.~Huang, L.~Liu, and X.~Xu,
\newblock ``Clothes collocation recommendations by compatibility learning,''
\newblock in {\em IEEE International Conference on Web Services}, 2018, pp.
  179--186.

\bibitem{Li17}
Y.~Li, L.~Cao, J.~Zhu, and J.~Luo,
\newblock ``Mining fashion outfit composition using an end-to-end deep learning
  approach on set data,''
\newblock {\em IEEE Transactions on Multimedia}, vol. 19, no. 8, pp.
  1946--1955, 2017.

\bibitem{Han17}
X.~Han, Z.~Wu, Y.~Jiang, and L.S. Davis,
\newblock ``Learning fashion compatibility with bidirectional {LSTM}s,''
\newblock in {\em ACM on Multimedia Conference}, 2017, pp. 1078--1086.

\bibitem{Vasi18}
M.I. Vasileva, B.A. Plummer, K.~Dusad, S.~Rajpal, R.~Kumar, and D.~Forsyth,
\newblock ``Learning type-aware embeddings for fashion compatibility,''
\newblock {\em arXiv preprint arXiv:1803.09196}, 2018.

\bibitem{Veit15}
A.~Veit, B.~Kovacs, S.~Bell, J.~McAuley, K.~Bala, and S.~Belongie,
\newblock ``Learning visual clothing style with heterogeneous dyadic
  co-occurrences,''
\newblock in {\em IEEE International Conference on Computer Vision}, 2015, pp.
  4642--4650.

\bibitem{Jaga14}
V.~Jagadeesh, R.~Piramuthu, A.~Bhardwaj, W.~Di, and N.~Sundaresan,
\newblock ``Large scale visual recommendations from street fashion images,''
\newblock in {\em ACM SIGKDD International Conference on Knowledge discovery
  and data mining}, 2014, pp. 1925--1934.

\bibitem{Zhan15}
Z.~Zhang, P.~Luo, C.~Loy, and X.~Tang,
\newblock ``Learning social relation traits from face images,''
\newblock in {\em IEEE International Conference on Computer Vision}, 2015, pp.
  3631--3639.

\bibitem{Simo14}
K.~Simonyan and A.~Zisserman,
\newblock ``Very deep convolutional networks for large-scale image
  recognition,''
\newblock {\em arXiv preprint arXiv:1409.1556}, 2014.

\bibitem{Heo11}
Y.S. Heo, K.M. Lee, and S.U. Lee,
\newblock ``Robust stereo matching using adaptive normalized
  cross-correlation,''
\newblock {\em IEEE Transactions on Pattern Analysis and Machine Intelligence},
  vol. 33, no. 4, pp. 807--822, 2011.

\bibitem{Zhan12}
Yu~Zhang and Dit-Yan Yeung,
\newblock ``A convex formulation for learning task relationships in multi-task
  learning,''
\newblock {\em arXiv preprint arXiv:1203.3536}, 2012.

\bibitem{King14}
D.P. Kingma and J.~Ba,
\newblock ``Adam: A method for stochastic optimization,''
\newblock {\em arXiv preprint arXiv:1412.6980}, 2014.

\end{thebibliography}

\end{document}